\documentclass[10pt,twocolumn,letterpaper]{article}

\usepackage{amsmath}
\usepackage{graphicx}
\usepackage[table,xcdraw]{xcolor}
\usepackage{multirow}
\usepackage{subcaption} 
\usepackage{diagbox}
\usepackage{tabularx}
\usepackage{booktabs}
\usepackage{makecell}

\usepackage{iccv}              

\definecolor{iccvblue}{rgb}{0.21,0.49,0.74}
\usepackage[pagebackref,breaklinks,colorlinks,allcolors=iccvblue]{hyperref}

\def\paperID{7310} 
\def\confName{ICCV}
\def\confYear{2025}

\title{Revisiting Adversarial Patch Defenses on Object Detectors:\\Unified Evaluation, Large-Scale Dataset, and New Insights}

\author {
    Junhao Zheng$ ^1$$\quad$
    Jiahao Sun$ ^1$$\quad$
    Chenhao Lin$ ^1$\footnotemark[2]$\quad$
    Zhengyu Zhao$ ^1$\footnotemark[2]$\quad$
    Chen Ma$^1$$\quad$\\
    Chong Zhang$^1$$\quad$
    Cong Wang$^2$$\quad$
    Qian Wang$^3$$\quad$
    Chao Shen$ ^1$\\
    $^1 $\textit{Xi'an Jiaotong University}$\quad$
    $^2 $\textit{City University of Hong Kong}$\quad$
    $^3 $\textit{Wuhan University}
}

\begin{document}


\maketitle
\footnotetext[2]{Corresponding authors.}

\begin{abstract}
Developing reliable defenses against patch attacks on object detectors has attracted increasing interest.
However, we identify that existing defense evaluations lack a unified and comprehensive framework, resulting in inconsistent and incomplete assessments of current methods.
To address this issue, we revisit 11 representative defenses and present the first patch defense benchmark, involving 2 attack goals, 13 patch attacks, 11 object detectors, and 4 diverse metrics.
This leads to the large-scale adversarial patch dataset with 94 types of patches and 94,000 images. 
Our comprehensive analyses reveal new insights: (1) The difficulty in defending against naturalistic patches lies in the data distribution, rather than the commonly believed high frequencies.
Our new dataset with diverse patch distributions can be used to improve existing defenses by 15.09\% AP@0.5.
(2) The average precision of the attacked object, rather than the commonly pursued patch detection accuracy, shows high consistency with defense performance.
(3) Adaptive attacks can substantially bypass existing defenses, and defenses with complex/stochastic models or universal patch properties are relatively robust.
We hope that our analyses will serve as guidance on properly evaluating patch attacks/defenses and advancing their design.
Code and dataset are available at \href{https://github.com/Gandolfczjh/APDE}{https://github.com/Gandolfczjh/APDE}, where we will keep integrating new attacks/defenses.

\end{abstract}    
\section{Introduction}
\label{sec:intro}

Deep neural networks (DNNs) have been widely applied across various real-world domains, including pedestrian object detection~\cite{yolov2, ddetr}, depth estimation~\cite{monodepth2}, and autonomous driving~\cite{Autonomous}. Despite their outstanding performance, DNNs are known to be vulnerable to adversarial attacks, which severely limits their reliability~\cite{zhao2020towards, zhao2021success}. These attacks manipulate input data, causing DNNs to produce incorrect predictions, potentially leading to severe incidents. Adversarial attacks can be realized in different forms, including adversarial perturbations~\cite{Quantization} and adversarial patches~\cite{3D2Fool, SlowTrack, controlloc}. Among these, adversarial patch attacks, which are feasible in the physical world, pose a significant threat to DNNs in real-world applications.

In recent years, substantial efforts have been made to defend against adversarial patches~\cite{LGS,NAPGuard,Zmask,Diffender,SAC,NutNet}. For instance, LGS~\cite{LGS} assumes that adversarial patches exhibit more intense pixel variations than clean images, aiming to detect and then smooth regions with significant texture changes to eliminate patches. Defense methods like PAD~\cite{PAD} and NAPGuard~\cite{NAPGuard} utilize additional segmentation~\cite{U-Net} or detection~\cite{yolov2,yolov3} models to eliminate adversarial patches. Approaches like DIFFender~\cite{Diffender} and NutNet~\cite{NutNet} leverage differences in data distribution between adversarial and clean images, using generative models~\cite{diffusion,GAN} to analyze and locate adversarial patches.

However, so far, there still lacks a unified framework for defense evaluation, and as a result, the effectiveness of various defenses may not be assessed consistently and comprehensively.
Specifically, we identify the main limitations of existing evaluations and address them through the following four aspects:
(1) \textbf{Unified framework.} Existing studies suffer from inconsistent parameters, attack methods, and patch placement strategies, which impede fair comparisons. To address this, we unify the evaluation paradigm around pedestrian detection, standardizing parameters, and patch deployment.
(2) \textbf{Suitable metrics.} Some works use only patch detection accuracy to assess defenses~\cite{Adyolo, NAPGuard}, thereby raising fairness concerns because high detection accuracy does not necessarily reflect superior defense performance. To address this, we adopt average precision (AP) and attack success rate (ASR) as comprehensive evaluation metrics and experimentally demonstrate that patch detection accuracy is unsuitable for objectively comparing the defense performance.
(3) \textbf{Comprehensive analyses.} Existing studies frequently overlook critical factors such as real-time performance, defense against patches of varying sizes and types, and real-world applicability. To address this, we systematically analyze defense efficiency and performance across different patch sizes, attack types, adaptive attacks, and patch mask filling strategies, while extending evaluations to the physical world.
(4) \textbf{Challenging attacks.} Existing studies utilize adversarial patch datasets (e.g., GAP~\cite{NAPGuard} and Apricot~\cite{Apricot}) that lack categorization by detector type and comprehensive attack coverage, making them insufficient for evaluating defense robustness. To address this, we conduct experiments under white-box attack conditions, offering a worst-case analysis grounded in data distribution to rigorously assess defense performance.

Overall, we provide the first patch defense benchmark on evaluating patch defenses and introduce a new large-scale Adversarial Patch Defense Evaluation (APDE) dataset, containing 94 types of patches and 94,000 images.
In comparison, popular datasets in existing work are far smaller: Apricot (60 types of patches, 1,011 images)~\cite{Apricot} and GAP (25 types of patches, 9,266 images)~\cite{NAPGuard}.
Our main contributions can be summarized as follows:

\begin{itemize}
\item We identify four main limitations of existing evaluations of adversarial patch defenses, and we provide the first patch defense benchmark, involving 11 representative defense methods, 2 attack goals, 13 patch attacks, 11 object detectors, and 4 diverse metrics.
\item We conduct comprehensive analyses to reveal new insights, including the root causes of defense failures against naturalistic patches, unfair comparative metrics in prior defense performance evaluations, and effective strategies to counter adaptive attacks.
\item We construct the large-scale Adversarial Patch Defense Evaluation (APDE) dataset.
Beyond evaluating defenses, it can also be used to substantially improve defenses, with an average improvement of 15.09\% AP@0.5 on mainstream defense models.
\end{itemize}
\section{Related work}
\label{sec:related}

\subsection{Object Detection}

Object detection is a core technology in the field of computer vision, aimed at identifying one or more predefined objects in images or videos and accurately locating their positions. Based on the distinction between detection box proposals and object classification, object detection is primarily divided into two categories: single-stage detectors (such as YOLO~\cite{yolov2,yolov3,yolov4,yolov5,yolov7}, SSD~\cite{SSD}, RetinaNet~\cite{retinanet}, CenterNet~\cite{centernet}, DDETR~\cite{ddetr}) and two-stage detectors (such as FRCNN~\cite{fasterrcnn}, MRCNN~\cite{maskrcnn}). Single-stage detectors directly predict the class probabilities and detection boxes of objects in an image through a single network, achieving both localization and classification simultaneously. In contrast, two-stage detectors first generate a set of region proposals in the image and then classify these proposals as objects of various categories or as background, typically employing separate networks or modules for processing.

\subsection{Adversarial Attacks on Object Detection}

Adversarial attacks on object detection add perturbations to input images, causing detection boxes to disappear, appear incorrectly, or label changes~\cite{Advpatch,chow2020adversarial}. Early work focuses on digital-domain attacks, adding small, global perturbations to image pixels~\cite{DAG}. To make attacks more practical, later research has shifted towards adversarial patches with large, local perturbations~\cite{LaVAN,Advpatch}. Adversarial patch attacks against object detectors place an optimized patch on an object.
Adversarial patches can execute either \textit{hiding attacks} or \textit{appearing attacks}. These attacks can be further classified into optimization-based and generator-based methods. Specifically, hiding attacks enable targeted objects to evade detection by the detector. Optimization-based attacks~\cite{T-SEA,Advcloak,AdvTshirt,Advpatch} begin with a noise patch and iteratively adjust pixel values to maximize detector error, often employing optimization techniques such as gradient descent to update the patch. Generator-based attacks~\cite{GNAP,DM-NAP,TC-EGA} use generative models~\cite{GAN,diffusion} to directly produce adversarial patches, offering the advantage of creating more diverse and natural-looking patches~\cite{zheng2024breaking}. Differently, appearing attacks~\cite{AA_patch} disguise patches to resemble normal target objects, causing the detector to misidentify them.

\subsection{Adversarial Defenses on Object Detection}

Defense methods against adversarial perturbations have been widely explored, including techniques such as image denoising~\cite{HGD} and scaling~\cite{xiao2019seeing}. However, these methods are less effective against adversarial patch attacks, as patches are unrestricted. In recent years, numerous defenses have emerged specifically targeting adversarial patches~\cite{Zmask,SAC,NAPGuard}. These defenses can be classified into \textbf{certified defenses} and \textbf{empirical defenses}. Certified defenses~\cite{detectorguard, objectseeker} require strict assumptions of the threat model and are limited by predefined patch size and quantity. In contrast, empirical defenses can be further categorized into three subtypes based on their underlying principles: defenses based on patch detection/segmentation~\cite{SAC,PAD,NAPGuard,Adyolo}, defenses leveraging prior knowledge of patches~\cite{Zmask,APE,Jedi,LGS}, and defenses based on generative models~\cite{Diffender,NutNet}. The overview of the defenses evaluated in this study is shown in Tab.~\ref{tab:categorized_defenses}. (Implementation details of the defense methods are provided in Appendix A.2)

\begin{table}[!t]
\centering
\small 
\begin{tabularx}{\columnwidth}{ X X }
    \hline
    \textbf{Method} & \textbf{Category} \\
    \hline
    SAC~\cite{SAC} (CVPR'22) & Patch Segmentation\\
    PAD~\cite{PAD} (CVPR'24) & Patch Segmentation \\
    Adyolo~\cite{Adyolo} (ArXiv'21) & Patch Detection \\
    NAPGuard~\cite{NAPGuard}(CVPR'24) & Patch Detection \\
    \hline
    LGS~\cite{LGS} (WACV'19) & Prior Knowledge of Patches \\
    Zmask~\cite{Zmask} (AAAI'23) & Prior Knowledge of Patches \\
    Jedi~\cite{Jedi} (CVPR'23) & Prior Knowledge of Patches \\
    \hline
    DIFFender~\cite{Diffender} (ECCV'24) & Generative Models \\
    NutNet~\cite{NutNet} (CCS'24) & Generative Models \\
    \hline
    \hline
    DetectorGuard~\cite{detectorguard}(CCS'21) & Certified Defenses \\
    Objectseeker~\cite{objectseeker} (S$\&$P'23) & Certified Defenses \\
    \hline
\end{tabularx}
\caption{Overview of our evaluated \textbf{Empirical/Certified} defenses.}
\label{tab:categorized_defenses}
\end{table}



\section{New Adversarial Patch Dataset}
\label{sec:addataset}

\subsection{Adversarial Patch Generation}
In this paper, we generate adversarial patches, denoted as $\delta$, by conducting white-box adversarial attacks on various object detectors. Let $\mathcal{F}=\{f_1,f_2,\ldots,f_n\}$ be the set of object detectors, where a single detector $f_i:X\longmapsto Y$ takes input images $x \in X$ and label $y \in Y$. We apply the adversarial patch $\delta$ to the input image $x$ through an applier $\mathcal{A}$ with transformations $t \in T$ (e.g., rotation, scaling, etc.). In this work, we focus on universal adversarial patches, which can be expressed as:
\begin{equation}
  \delta^* = \arg \min_{\delta} \mathbb{E}_{x \sim X} \left[ \mathcal{L}(f_i(\mathcal{A}(x, \delta, t)), y) \right] + \lambda L_{tv}(\delta)
  \label{eq:equation1}
\end{equation}
where the loss function $\mathcal{L}(\cdot)$ measures the corruption of the detector, typically using object or class confidence, and $L_{tv}(\cdot)$ is the total variation loss, which encourages the generated adversarial patch to be smoother.

Different attack methods utilize various loss functions and strategies for optimizing the adversarial patches, resulting in multiple optimization spaces and differences in the generated patches. Some patches exhibit better attack performance, while others are smoother and more natural. A diverse set of patches is necessary for a comprehensive evaluation of defense methods.


\subsection{Adversarial Dataset Construction}
\label{sec:PDE_dataset}
Based on the above-described principles for adversarial patch generation, we construct the large-scale Adversarial Patch Defense Evaluation (APDE) dataset. According to statistics, INRIA-Person~\cite{inria} and MS COCO~\cite{MSCOCO} are the main (clean) datasets used for adversarial attacks and defenses against object detection. To ensure diversity, we select these two datasets to generate 94 types of adversarial patches using 13 adversarial patch attack methods and 11 detectors.
Implementation details of these attack methods are provided in Appendix A.1.

Each patch is applied to the test sets of INRIA-Person and MS COCO, resulting in 94,000 images in total. All images are stored in PNG format with a fixed size of 416 × 416 pixels, achieved through padding or resizing, aligning with the settings described in the respective papers. The dataset is divided into a training set (56,400 images) and a testing set (37,600 images), following a 6:4 ratio.

Compared to existing adversarial patch datasets, the APDE dataset exhibits three prominent advantages:

\begin{itemize}
\item \textbf{Large Scale.} Compared to Apricot (1,011 images) and GAP (9,266 images), the APDE dataset contains 94,000 images, comprehensively covering various scenarios and attack types. This large-scale dataset enables a more accurate evaluation of model generalization and robustness, providing an unprecedented data foundation for in-depth research on patch defense. 
\item \textbf{Diverse Patches.} Compared to Apricot (60 types) and GAP (25 types), the APDE dataset includes 94 types of patches. These patches come in various shapes, including common geometric forms such as circles, squares, and rectangles, as well as natural shapes like dogs or cartoon patterns. The APDE dataset offers a diverse data distribution for training and evaluating defense models, enabling a more thorough evaluation of patch defenses.
\item \textbf{White-Box Setting.} Adversarial patches in the APDE dataset are trained on 11 mainstream object detectors, creating a white-box evaluation setting for patch defenses. This setup poses the worst-case adversarial threat to the detectors, allowing for the measurement of a defense model's worst-case performance and the identification of weaknesses in various defense methods.
\end{itemize}

The above advantages also make our APDE dataset more effective than previous patch datasets~\cite{Apricot, NAPGuard}, in training robust defense models, even for out-of-domain attacks that are unseen during training.
See detailed experimental results in Section~\ref{sec:retrain} and Appendix B.3.



\section{Experiments}
\label{sec:eval}


\begin{figure*}[!t]
  \centering
  \includegraphics[width=\textwidth]{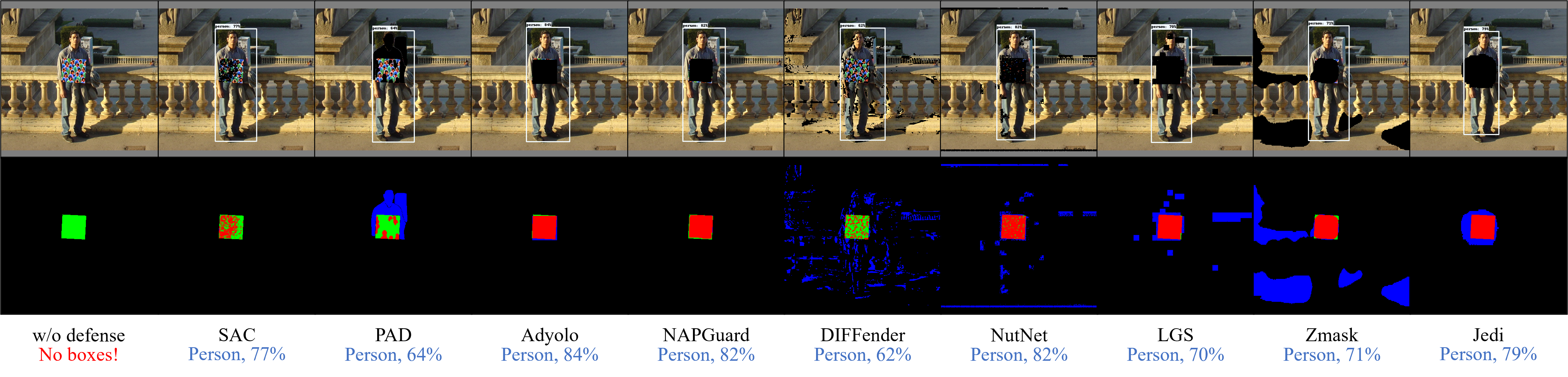}
  \caption{\textbf{Comparison of patch detection and defense performance.} \textbf{The first row}: Images with adversarial patches. \textbf{The second row}: Patch mask images generated by defenses (\textcolor{green}{Green}: the true location of patches. \textcolor{red}{Red}: detected patches by the defense. \textcolor{blue}{Blue}: the background mistakenly identified as patches). Without defenses, the adversarial patch causes an incorrect prediction (in the first column), while defenses correct the prediction (in the remaining columns).}
  \label{fig:main_defense_effect}
  \hfill
\end{figure*}

In this section, we first outline the evaluation metrics (Sec~\ref{sec:metrics}) and then systematically evaluate defense methods. We investigate the correlation between patch detection accuracy and actual defense performance in digital domains (Sec~\ref{sec:performance}), validate the practicality of existing defenses in physical-world scenarios (Sec~\ref{sec:physical}), and analyze the defense robustness against adaptive attacks (Sec~\ref{sec:adaptive}). Finally, we conduct an empirical evaluation of certified defenses under constrained threat models (Sec~\ref{sec:certified}).


\subsection{Evaluation Metrics}
\label{sec:metrics}
Some existing studies, however, rely solely on patch detection accuracy (e.g., patch AP@0.5) as the primary criterion for defense effectiveness~\cite{Adyolo, NAPGuard}, raising fairness concerns. To address this limitation, we use widely adopted \textbf{average precision (AP)} and \textbf{attack success rate (ASR)} as core metrics, while experimentally demonstrating the inadequacy of patch detection accuracy for objective comparisons (Sec~\ref{sec:performance}).
To further enhance evaluation rigor, we introduce two metrics:
\textbf{Defense efficiency}: We measure the inference time per image (in milliseconds) to assess computational practicality.
\textbf{Mask-based detection accuracy}: We adopt mean Intersection over Union (mIoU)~\cite{NmIoU} instead of patch AP@0.5. This adjustment is critical because irregular or non-rectangular masks produced by certain defense methods cannot be accurately evaluated via axis-aligned bounding boxes. The mIoU metric directly quantifies spatial overlap between predicted and ground-truth masks, providing a shape-agnostic measure of localization precision.
\begin{align} 
\text{SmIoU} =& \frac{\sum_{n=1}^{N} \text{TP}_{n}}{\sum_{n=1}^{N} (\text{FP}_{n} + \text{TP}_{n} + \text{FN}_{n})}\\
\text{NmIoU} =& \frac{1}{N} \sum_{n=1}^{N} \frac{\text{TP}_{n}}{\text{FP}_{n} + \text{TP}_{n} + \text{FN}_{n}}
\end{align}
SmIoU first aggregates predictions for positive and negative samples across all images, making it more sensitive to large pixel areas. NmIoU, on the other hand, averages the mIoU results of each image individually, reflecting image-level performance.

\renewcommand{\arraystretch}{1.25}  
\begin{table*}[ht]
\centering
\resizebox{\textwidth}{!}{
\begin{tabular}{c | cc | cc | cc | cc | cc | cc | cc | cc | cc }
\hline
\multirow{2}{*}{\textbf{Model} (w/o defense)} & \multicolumn{2}{c}{\textbf{SAC}~\cite{SAC}} & \multicolumn{2}{c}{\textbf{PAD}~\cite{PAD}} & \multicolumn{2}{c}{\textbf{Adyolo}~\cite{Adyolo}} & \multicolumn{2}{c}{\textbf{NAPGuard}\cite{NAPGuard}} & \multicolumn{2}{c}{\textbf{DIFFender}\cite{Diffender}} & \multicolumn{2}{c}{\textbf{NutNet}~\cite{NutNet}} & \multicolumn{2}{c}{\textbf{LGS}~\cite{LGS}} & \multicolumn{2}{c}{\textbf{Zmask}~\cite{Zmask}} & \multicolumn{2}{c}{\textbf{Jedi}~\cite{Jedi}} \\ \cline{2-19}
 & mean & min & mean & min & mean & min & mean & min & mean & min & mean & min & mean & min & mean & min & mean & min \\ \hline
\textbf{YOLOv2} (21.73) & \cellcolor[HTML]{d8d8d8}52.08 & \cellcolor[HTML]{dfdfdf}29.23 & \cellcolor[HTML]{bcbcbc}\underline{67.22} & \cellcolor[HTML]{b9b9b9}\underline{56.48} & \cellcolor[HTML]{cfcfcf}57.01 & \cellcolor[HTML]{c5c5c5}48.10 & \cellcolor[HTML]{c1c1c1}64.78 & \cellcolor[HTML]{c6c6c6}47.48 & \cellcolor[HTML]{d9d9d9}51.35 & \cellcolor[HTML]{d2d2d2}38.38 & \cellcolor[HTML]{bababa}\textbf{68.68} & \cellcolor[HTML]{aeaeae}\textbf{65.03} & \cellcolor[HTML]{c4c4c4}63.04 & \cellcolor[HTML]{c4c4c4}48.67 & \cellcolor[HTML]{dcdcdc}49.46 & \cellcolor[HTML]{dddddd}31.02 & \cellcolor[HTML]{d5d5d5}53.31 & \cellcolor[HTML]{c8c8c8}45.66 \\ 

\textbf{YOLOv3} (28.4) & \cellcolor[HTML]{b2b2b2}72.92 & \cellcolor[HTML]{c4c4c4}48.89 & \cellcolor[HTML]{999999}\underline{86.65} & \cellcolor[HTML]{9d9d9d}76.85 & \cellcolor[HTML]{b0b0b0}73.77 & \cellcolor[HTML]{b8b8b8}57.35 & \cellcolor[HTML]{969696}\textbf{88.50} & \cellcolor[HTML]{969696}\textbf{82.58} & \cellcolor[HTML]{b9b9b9}69.15 & \cellcolor[HTML]{c4c4c4}48.68 & \cellcolor[HTML]{9b9b9b}85.42 & \cellcolor[HTML]{979797}\underline{81.18} & \cellcolor[HTML]{9c9c9c}85.12 & \cellcolor[HTML]{b2b2b2}61.68 & \cellcolor[HTML]{aaaaaa}77.26 & \cellcolor[HTML]{b1b1b1}62.79 & \cellcolor[HTML]{b9b9b9}68.77 & \cellcolor[HTML]{b7b7b7}58.63 \\

\textbf{YOLOv4} (39.97) & \cellcolor[HTML]{b8b8b8}69.35 & \cellcolor[HTML]{cccccc}43.20 & \cellcolor[HTML]{adadad}75.64 & \cellcolor[HTML]{adadad}65.66 & \cellcolor[HTML]{b2b2b2}72.84 & \cellcolor[HTML]{b4b4b4}60.26 & \cellcolor[HTML]{999999}\textbf{86.59} & \cellcolor[HTML]{969696}\textbf{81.92} & \cellcolor[HTML]{c4c4c4}62.98 & \cellcolor[HTML]{d3d3d3}38.17 & \cellcolor[HTML]{a0a0a0}\underline{82.80} & \cellcolor[HTML]{9f9f9f}\underline{75.66} & \cellcolor[HTML]{b0b0b0}73.78 & \cellcolor[HTML]{c4c4c4}48.53 & \cellcolor[HTML]{c6c6c6}61.73 & \cellcolor[HTML]{d2d2d2}38.54 & \cellcolor[HTML]{c6c6c6}61.71 & \cellcolor[HTML]{c1c1c1}51.05 \\ 

\textbf{YOLOv5} (46.03) & \cellcolor[HTML]{c0c0c0}64.85 & \cellcolor[HTML]{c5c5c5}48.16 & \cellcolor[HTML]{a1a1a1}\underline{82.27} & \cellcolor[HTML]{9d9d9d}\textbf{76.93} & \cellcolor[HTML]{b6b6b6}70.57 & \cellcolor[HTML]{b9b9b9}57.04 & \cellcolor[HTML]{9e9e9e}\textbf{83.88} & \cellcolor[HTML]{a1a1a1}74.06 & \cellcolor[HTML]{c7c7c7}61.02 & \cellcolor[HTML]{cdcdcd}42.03 & \cellcolor[HTML]{a3a3a3}80.92 & \cellcolor[HTML]{a0a0a0}\underline{74.83} & \cellcolor[HTML]{b3b3b3}72.45 & \cellcolor[HTML]{c0c0c0}51.60 & \cellcolor[HTML]{c4c4c4}62.72 & \cellcolor[HTML]{cbcbcb}43.74 & \cellcolor[HTML]{c1c1c1}64.53 & \cellcolor[HTML]{c2c2c2}50.13 \\ 

\textbf{YOLOv7} (65.14) & \cellcolor[HTML]{b0b0b0}74.19 & \cellcolor[HTML]{a9a9a9}68.41 & \cellcolor[HTML]{acacac}\underline{75.95} & \cellcolor[HTML]{a1a1a1}\textbf{73.90} & \cellcolor[HTML]{b1b1b1}73.52 & \cellcolor[HTML]{adadad}65.27 & \cellcolor[HTML]{a9a9a9}\textbf{77.74} & \cellcolor[HTML]{a6a6a6}\underline{70.48} & \cellcolor[HTML]{d0d0d0}56.53 & \cellcolor[HTML]{bebebe}53.07 & \cellcolor[HTML]{c3c3c3}63.66 & \cellcolor[HTML]{b6b6b6}59.34 & \cellcolor[HTML]{b2b2b2}72.71 & \cellcolor[HTML]{acacac}66.18 & \cellcolor[HTML]{b8b8b8}69.26 & \cellcolor[HTML]{b2b2b2}61.76 & \cellcolor[HTML]{b7b7b7}69.99 & \cellcolor[HTML]{adadad}65.79 \\ 

\textbf{SSD} (16.59) & \cellcolor[HTML]{e1e1e1}46.77 & \cellcolor[HTML]{e7e7e7}23.69 & \cellcolor[HTML]{b9b9b9}\textbf{69.23} & \cellcolor[HTML]{b2b2b2}\underline{61.95} & \cellcolor[HTML]{dedede}48.45 & \cellcolor[HTML]{d4d4d4}37.41 & \cellcolor[HTML]{bfbfbf}65.48 & \cellcolor[HTML]{c6c6c6}47.68 & \cellcolor[HTML]{dadada}50.82 & \cellcolor[HTML]{d1d1d1}39.71 & \cellcolor[HTML]{b9b9b9}\underline{69.20} & \cellcolor[HTML]{acacac}\textbf{66.33} & \cellcolor[HTML]{c0c0c0}64.91 & \cellcolor[HTML]{c3c3c3}49.36 & \cellcolor[HTML]{cfcfcf}56.65 & \cellcolor[HTML]{cccccc}42.83 & \cellcolor[HTML]{e0e0e0}47.51 & \cellcolor[HTML]{cfcfcf}40.93 \\ 

\textbf{CenterNet} (14.66) & \cellcolor[HTML]{dfdfdf}47.83 & \cellcolor[HTML]{e8e8e8}22.39 & \cellcolor[HTML]{b9b9b9}\underline{69.03} & \cellcolor[HTML]{cfcfcf}40.99 & \cellcolor[HTML]{dbdbdb}50.29 & \cellcolor[HTML]{d3d3d3}37.74 & \cellcolor[HTML]{c2c2c2}64.00 & \cellcolor[HTML]{bfbfbf}\underline{52.36} & \cellcolor[HTML]{f1f1f1}38.39 & \cellcolor[HTML]{efefef}17.47 & \cellcolor[HTML]{b5b5b5}\textbf{71.04} & \cellcolor[HTML]{b1b1b1}\textbf{62.71} & \cellcolor[HTML]{c6c6c6}61.85 & \cellcolor[HTML]{c8c8c8}45.69 & \cellcolor[HTML]{e4e4e4}45.53 & \cellcolor[HTML]{ededed}18.79 & \cellcolor[HTML]{dddddd}49.43 & \cellcolor[HTML]{cfcfcf}40.82 \\ 

\textbf{RetinaNet} (25.84) & \cellcolor[HTML]{c5c5c5}62.20 & \cellcolor[HTML]{d1d1d1}39.30 & \cellcolor[HTML]{a3a3a3}\underline{81.26} & \cellcolor[HTML]{9a9a9a}\textbf{79.37} & \cellcolor[HTML]{c4c4c4}63.09 & \cellcolor[HTML]{cdcdcd}42.29 & \cellcolor[HTML]{acacac}76.36 & \cellcolor[HTML]{b6b6b6}58.76 & \cellcolor[HTML]{d5d5d5}53.51 & \cellcolor[HTML]{d1d1d1}39.69 & \cellcolor[HTML]{a0a0a0}\textbf{82.75} & \cellcolor[HTML]{a5a5a5}\underline{71.18} & \cellcolor[HTML]{adadad}75.78 & \cellcolor[HTML]{cccccc}42.87 & \cellcolor[HTML]{dadada}51.00 & \cellcolor[HTML]{e5e5e5}24.81 & \cellcolor[HTML]{cbcbcb}59.31 & \cellcolor[HTML]{c8c8c8}45.98 \\ 

\textbf{MRCNN} (28.89) & \cellcolor[HTML]{bcbcbc}67.42 & \cellcolor[HTML]{b6b6b6}58.88 & \cellcolor[HTML]{9f9f9f}83.01 & \cellcolor[HTML]{a3a3a3}73.11 & \cellcolor[HTML]{b8b8b8}69.67 & \cellcolor[HTML]{d3d3d3}38.15 & \cellcolor[HTML]{9d9d9d}\underline{84.48} & \cellcolor[HTML]{a2a2a2}\underline{73.53} & \cellcolor[HTML]{b9b9b9}69.09 & \cellcolor[HTML]{b1b1b1}62.79 & \cellcolor[HTML]{999999}\textbf{86.80} & \cellcolor[HTML]{9f9f9f}\textbf{75.54} & \cellcolor[HTML]{a7a7a7}78.68 & \cellcolor[HTML]{b0b0b0}63.13 & \cellcolor[HTML]{cecece}57.23 & \cellcolor[HTML]{e2e2e2}27.10 & \cellcolor[HTML]{bdbdbd}66.93 & \cellcolor[HTML]{b9b9b9}57.07 \\ 

\textbf{FRCNN} (46.37) & \cellcolor[HTML]{bfbfbf}65.78 & \cellcolor[HTML]{bdbdbd}54.10 & \cellcolor[HTML]{a5a5a5}80.14 & \cellcolor[HTML]{aaaaaa}67.74 & \cellcolor[HTML]{b2b2b2}72.77 & \cellcolor[HTML]{b5b5b5}60.04 & \cellcolor[HTML]{9a9a9a}\textbf{85.81} & \cellcolor[HTML]{9f9f9f}\textbf{75.52} & \cellcolor[HTML]{b9b9b9}68.87 & \cellcolor[HTML]{bababa}56.34 & \cellcolor[HTML]{9e9e9e}\underline{83.95} & \cellcolor[HTML]{a0a0a0}\underline{75.12} & \cellcolor[HTML]{a4a4a4}80.55 & \cellcolor[HTML]{bababa}55.89 & \cellcolor[HTML]{c5c5c5}62.25 & \cellcolor[HTML]{c9c9c9}44.87 & \cellcolor[HTML]{bbbbbb}68.08 & \cellcolor[HTML]{b8b8b8}57.66 \\ 

\textbf{DDETR} (4.56) & \cellcolor[HTML]{e2e2e2}46.33 & \cellcolor[HTML]{ebebeb}20.62 & \cellcolor[HTML]{bdbdbd}\textbf{66.88} & \cellcolor[HTML]{bdbdbd}\underline{53.67} & \cellcolor[HTML]{e9e9e9}42.38 & \cellcolor[HTML]{e3e3e3}26.59 & \cellcolor[HTML]{cdcdcd}57.75 & \cellcolor[HTML]{c7c7c7}46.93 & \cellcolor[HTML]{f4f4f4}36.78 & \cellcolor[HTML]{f7f7f7}11.98 & \cellcolor[HTML]{bdbdbd}\underline{66.56} & \cellcolor[HTML]{bababa}\textbf{55.79} & \cellcolor[HTML]{cccccc}58.52 & \cellcolor[HTML]{dfdfdf}29.55 & \cellcolor[HTML]{ffffff}30.73 & \cellcolor[HTML]{ffffff}6.43 & \cellcolor[HTML]{f2f2f2}37.75 & \cellcolor[HTML]{ededed}18.96 \\
\hline
\textbf{Overall} (30.74) & \cellcolor[HTML]{c8c8c8}60.88 & \cellcolor[HTML]{ebebeb}20.62 & \cellcolor[HTML]{acacac}\underline{76.12} & \cellcolor[HTML]{cfcfcf}40.99 & \cellcolor[HTML]{c4c4c4}63.12 & \cellcolor[HTML]{e3e3e3}26.59 & \cellcolor[HTML]{acacac}75.94 & \cellcolor[HTML]{c7c7c7}\underline{46.93} & \cellcolor[HTML]{d0d0d0}56.23 & \cellcolor[HTML]{f7f7f7}11.98 & \cellcolor[HTML]{ababab}\textbf{76.53} & \cellcolor[HTML]{bababa}\textbf{55.79} & \cellcolor[HTML]{b4b4b4}71.58 & \cellcolor[HTML]{dfdfdf}29.55 & \cellcolor[HTML]{cfcfcf}56.71 & \cellcolor[HTML]{ffffff}6.43 & \cellcolor[HTML]{cbcbcb}58.85 & \cellcolor[HTML]{ededed}18.96 \\ 
\hline
\textbf{Time Cost (ms)} & \multicolumn{2}{c|}{\textbf{44}}&  \multicolumn{2}{c|}{32100} & \multicolumn{2}{c|}{62} & \multicolumn{2}{c|}{\underline{59}} & \multicolumn{2}{c|}{1240} & \multicolumn{2}{c|}{71} & \multicolumn{2}{c|}{82} & \multicolumn{2}{c|}{417} & \multicolumn{2}{c}{349} \\ 
\hline

\end{tabular}
}
\caption{\textbf{Comparison of different defense performance against hiding attacks on 11 detectors.} We report mean and minimum person AP@0.5, as well as the time cost for each defense. SAC~\cite{SAC} exhibits the minimal time cost, while NutNet~\cite{NutNet} demonstrates the optimal defense performance. (In Appendix B.7, we provide detailed defense performance results against individual attacks.)} 
\label{tab:main_defense_perfomance}
\end{table*}

\subsection{Patch Defenses in the Digital Domain}
\label{sec:performance}
Fig.~\ref{fig:main_defense_effect} compares patch detection and defense performance. Tab.~\ref{tab:main_defense_perfomance} presents the mean and minimum person AP scores for each defense method against hiding attacks. Higher AP scores indicate better defense performance. (In addition to defenses against hiding attacks, we provide comprehensive experiments in supplementary material, including evaluations of defense method impacts on clean sample detection (Appendix B.1) and robustness assessments against appearing adversarial attacks (Appendix B.5))

\begin{figure}[!t]
    \centering
    \includegraphics[width=\columnwidth]{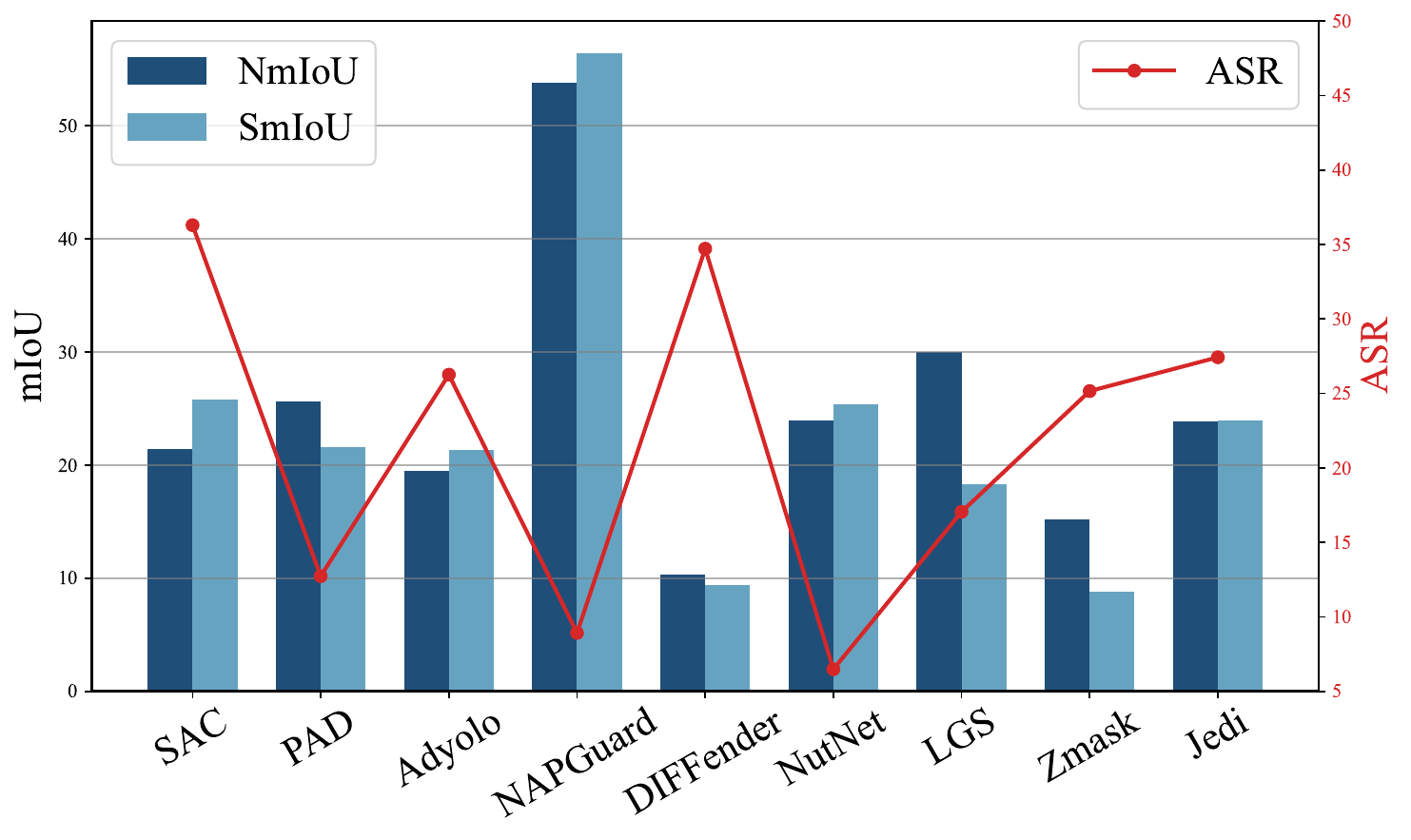}
    \caption{\textbf{Inconsistency in the relative strength of different defenses regarding patch detection accuracy vs. defense effectiveness.}
    For example, NAPGuard~\cite{NAPGuard} achieves the highest patch detection accuracy, while NutNet~\cite{NutNet} achieves the highest defense effectiveness.
    We report the NmIoU and SmIoU scores for patch detection, and the Attack Success Rate (ASR) after defense for defense effectiveness. 
}

    \label{fig:mIoU_asr}
\end{figure}

For the person AP@0.5, PAD~\cite{PAD}, NAPGuard~\cite{NAPGuard}, and NutNet~\cite{NutNet} significantly outperform others, dominating both mean and minimum AP results. Defense efficiency is also critical: as shown in Tab.~\ref{tab:main_defense_perfomance}, PAD's high time cost ($\sim$30s per image) limits its practicality. Defense performance varies widely across method categories, showing no systematic dependence on defense type. Among detectors, CenterNet~\cite{centernet} and DDETR~\cite{ddetr} have the lowest undefended AP scores, highlighting their need for robust adversarial defenses.
Notably, NutNet achieves the lowest ASR compared to other defenses, as visualized in Fig.~\ref{fig:mIoU_asr}. Overall, NutNet achieves the best defense performance and is also among the top in terms of defense efficiency.

Fig.~\ref{fig:mIoU_asr} illustrates the patch detection accuracy of each defense method. A higher mIoU score, the better the defense method’s generated mask aligns with the attack patch locations, thus indicating better detection performance. As shown in Fig.~\ref{fig:mIoU_asr}, the detection performance of NAPGuard~\cite{NAPGuard} significantly outperforms all other methods. However, NutNet~\cite{NutNet}, which achieves the best defense performance, shows only average detection ability. This finding underscores that strong patch detection does not necessarily equate to better defense performance. Therefore, \textbf{the average precision of the attacked object, rather than the commonly pursued patch detection accuracy, shows high consistency with defense performance.}

\begin{figure}[!t]
    \centering
    \includegraphics[width=\columnwidth]{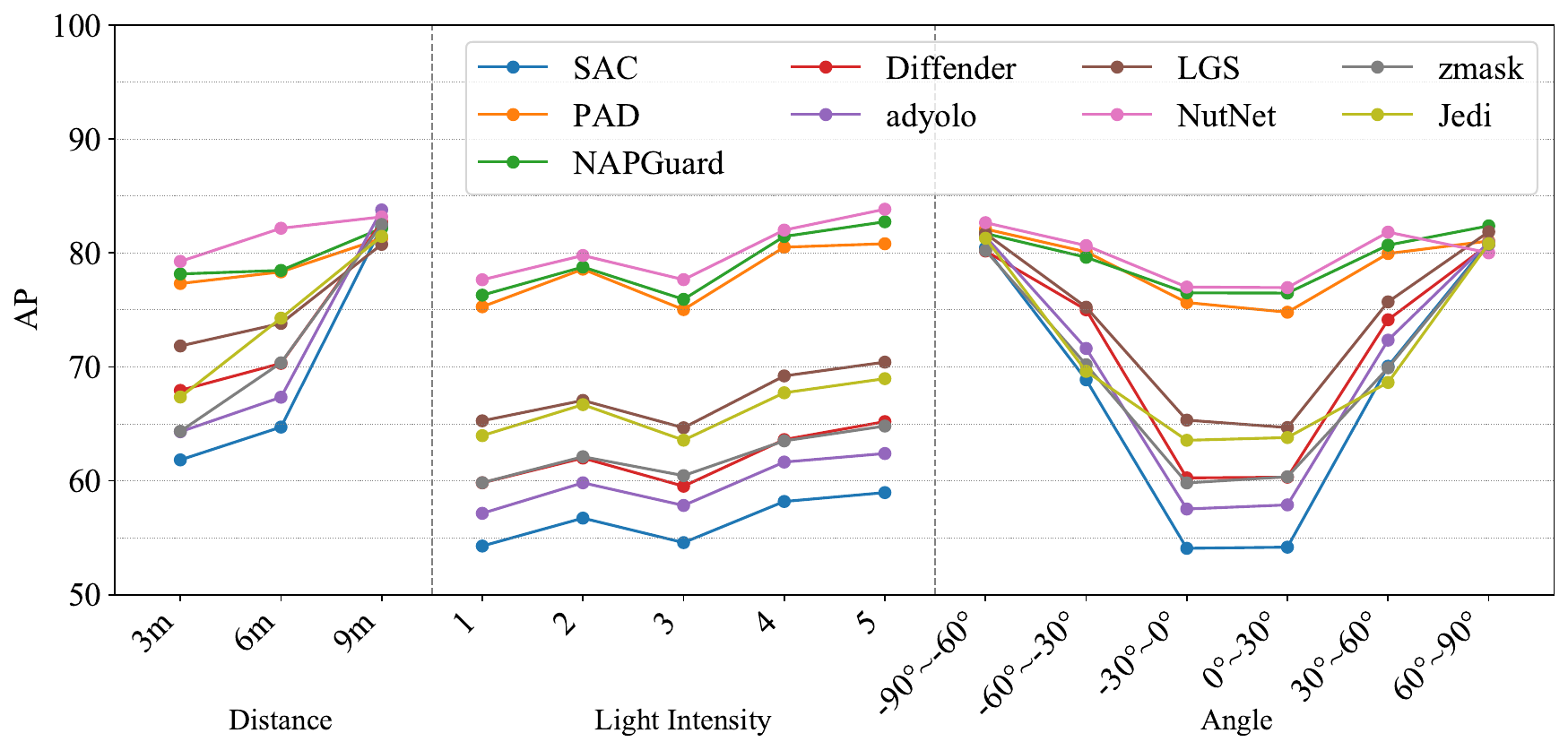}
    \caption{\textbf{Comparison of different defense methods in the physical world.} The defense performance is evaluated at different distances, light intensities, and angles.}
    \label{fig:physical1}
\end{figure}

\subsection{Patch Defenses in the Physical World}
\label{sec:physical}
\begin{figure}[!t]
    \centering
    \includegraphics[width=\columnwidth]{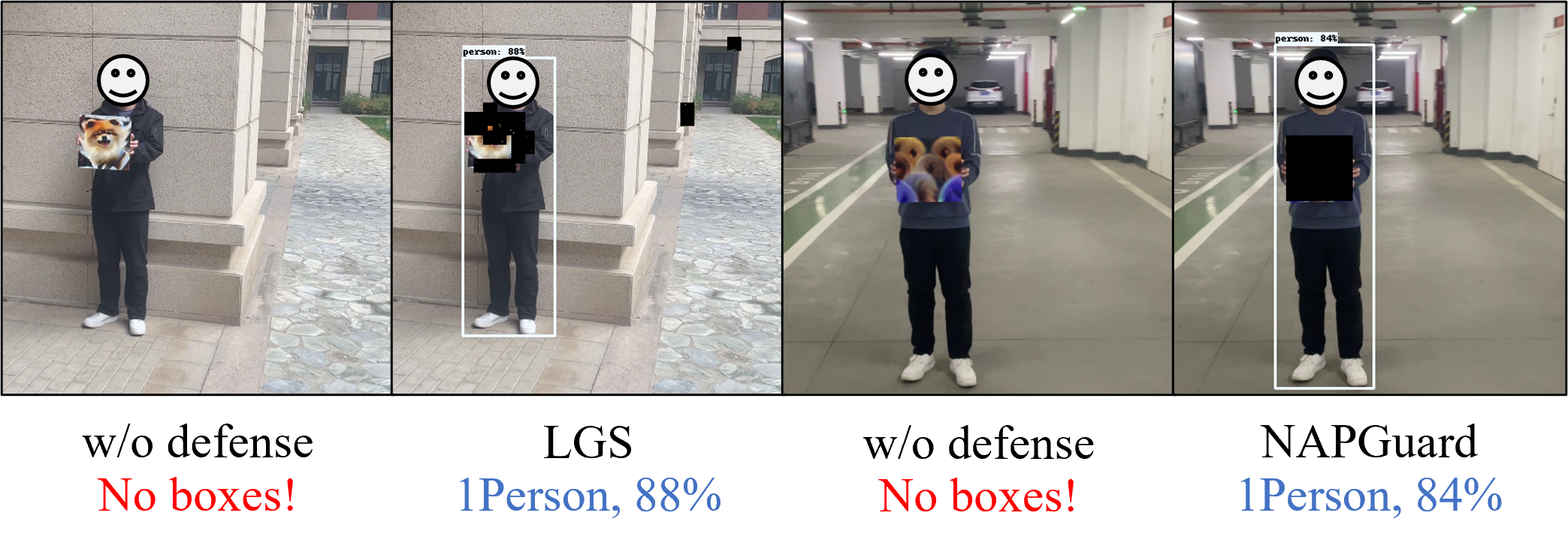}
    \caption{\textbf{Physical world attack and defense using a printed version of the patch.} The printed patch can successfully attack the detector, but it can be defended by LGS~\cite{LGS} and NAPGuard~\cite{NAPGuard}.}
    \label{fig:physical2}
\end{figure}

Adversarial patch attacks in the physical world can lead to serious security incidents. However, prior work has insufficiently explored defense performance in the physical world. Therefore, this section conducts experiments to investigate this issue. We select 30 types of patches and test them under different light intensities (1$\rightarrow$5, gradually increasing), at varying distances (3m/6m/9m), and angles (-90° to 90°). Using an iPhone16pro, we capture 540 frames of adversarial images. We test each defense and record the results, as shown in Fig.~\ref{fig:physical1}. 
Our experiments show that attack effectiveness declines with distance (e.g., AP rises from 60\% to 85\% as distance increases from 0m to 9m). Defenses perform worst in the [-30°, 30°) range, with AP scores 30-50\% lower than other angles. Attack effectiveness decreases while defense performance improves with increasing light intensity (levels 1-5).
As shown in Fig.~\ref{fig:physical2}, without defenses, the patch can successfully attack the victim detector under different angles, distances, and background environments. When using LGS~\cite{LGS} or NAPGuard~\cite{NAPGuard}, the patch is effectively eliminated. 
Therefore, defenses that perform well in the digital domain also generally demonstrate robustness in the physical world. Surprisingly, PAD~\cite{PAD}, which performs well in the digital domain, is surpassed by many other defense methods in the physical world.
Observing the patch masks generated by the defense methods, we find that although PAD can detect and remove attack patches, it is prone to misclassifying humans as patches, leading to incorrect predictions by the detector.

\begin{figure}[!t]
    \centering
    \includegraphics[width=\columnwidth]{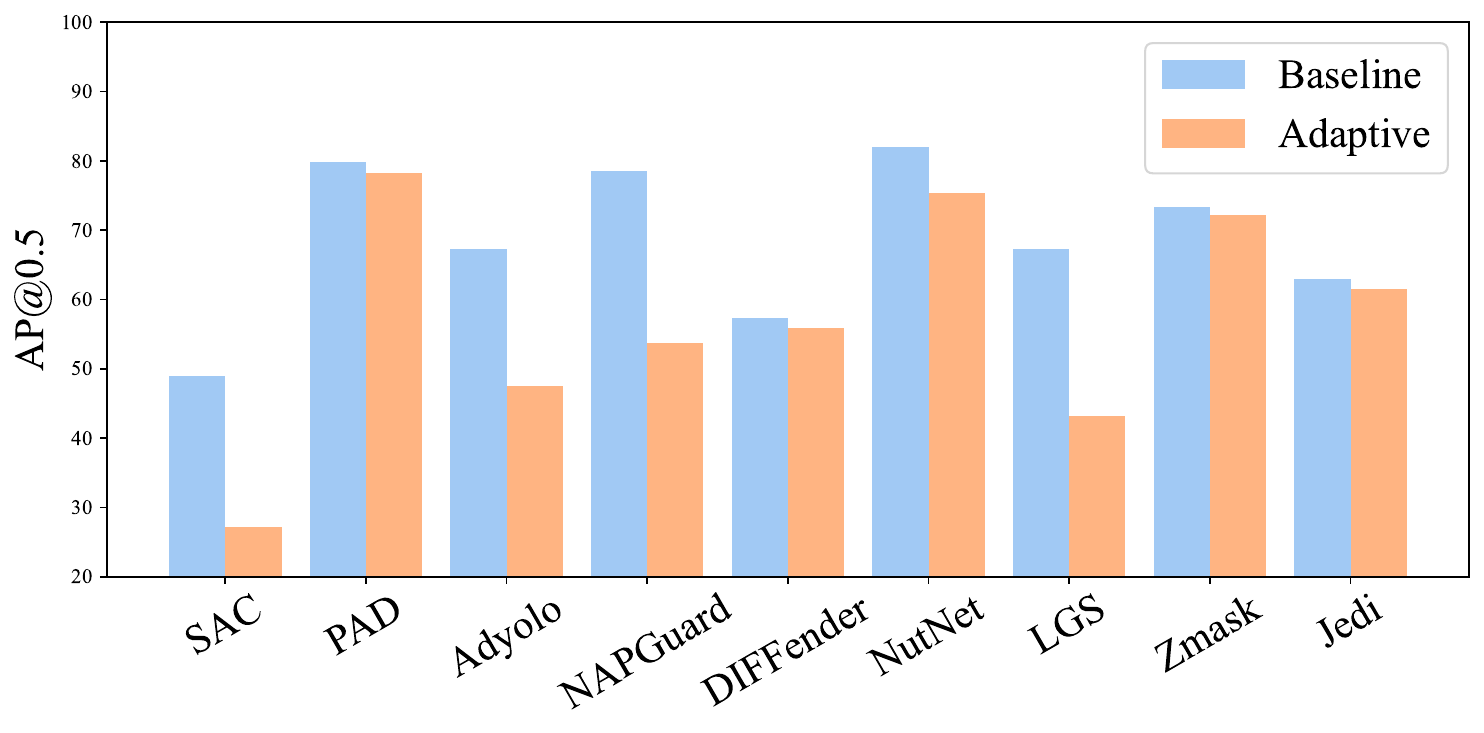}
    \caption{\textbf{Comparison of defense performance against baseline and adaptive attacks.} Adaptive attack resilience varies across defenses — PAD~\cite{PAD} remains robust, whereas SAC~\cite{SAC} is easily bypassed. (Details of the adaptive attacks are in Appendix A.3)}
    \label{fig:adaptive}
\end{figure}

\subsection{Patch Defenses against Adaptive Attacks}
\label{sec:adaptive}
An adaptive attack is designed with full knowledge of the defense mechanism, where adversaries specifically target the protection methods~\cite{Obfuscated}. The effectiveness against adaptive attacks is an important aspect in evaluating defense methods. While \cite{NutNet} argues that employing additional defense models to detect or filter adversarial patches may itself be vulnerable to adversarial attacks, and \cite{Diffender} suggests that some defense models are bypassed by adaptive attacks because their gradients can be easily obtained. Unlike empirical defenses, certified defenses theoretically prevent adaptive attacks within their threat models. As shown in Fig.~\ref{fig:adaptive}, empirical defenses like PAD remain relatively robust, whereas the others, such as SAC, are more vulnerable.

We analyze in detail the underlying principles of each defense method. For defenses based on patch detection or segmentation, only PAD remains largely unaffected. This is because PAD utilizes the complex SAM model~\cite{SAM} and relies on semantic differences, making it difficult to optimize gradients for adaptive attacks. For defenses based on generative models, DIFFender is less susceptible to adaptive attacks compared to NutNet, due to the inherent stochasticity of the Diffusion model~\cite{diffusion} used by DIFFender, which complicates the training of adaptive attacks. For defenses based on prior knowledge of patches, LGS is more easily bypassed by incorporating patch smoothness into the loss function, as pixel-level inconsistency occurs only in specific patches, lacking universality. Both Zmask and Jedi target universal patch properties: Zmask addresses feature-level over-activation, while Jedi tackles image-level high-entropy. Suppressing these characteristics inherently decreases the attack performance of patches, rendering adaptive attacks ineffective. In summary, \textbf{adaptive attacks can substantially bypass existing defenses, and defenses with complex/stochastic models or universal patch properties are relatively robust.}

\begin{table}[!t]
\resizebox{\columnwidth}{!}{
\begin{tabular}{c|ccc|ccc}
\hline
\multirow{2}{*}{\textbf{Model}} & \multicolumn{3}{c|}{\textbf{YOLOv3}} & \multicolumn{3}{c}{\textbf{FRCNN}} \\
\cline{2-7}
  & 1 & 2 & 3 & 1 & 2 & 3\\ \hline 
\textbf{w/o defense} & \cellcolor[HTML]{d8d8d8}32.20 & \cellcolor[HTML]{dbdbdb}29.83 & \cellcolor[HTML]{dedede}27.14 & \cellcolor[HTML]{cacaca}43.67 & \cellcolor[HTML]{cdcdcd}41.04 & \cellcolor[HTML]{d3d3d3}36.52 \\ \hline
\textbf{SAC}~\cite{SAC} & \cellcolor[HTML]{b4b4b4}61.92 & \cellcolor[HTML]{b8b8b8}58.51 & \cellcolor[HTML]{bababa}57.33 & \cellcolor[HTML]{aeaeae}67.35 & \cellcolor[HTML]{b0b0b0}65.62 & \cellcolor[HTML]{b5b5b5}61.39 \\
\textbf{PAD}~\cite{PAD} & \cellcolor[HTML]{979797}86.57 & \cellcolor[HTML]{9e9e9e}80.24 & \cellcolor[HTML]{9d9d9d}81.31 & \cellcolor[HTML]{999999}84.74 & \cellcolor[HTML]{9d9d9d}81.31 & \cellcolor[HTML]{9d9d9d}80.94 \\
\textbf{Adyolo}~\cite{Adyolo} & \cellcolor[HTML]{a9a9a9}71.08 & \cellcolor[HTML]{aeaeae}67.42 & \cellcolor[HTML]{b2b2b2}63.84 & \cellcolor[HTML]{a7a7a7}73.07 & \cellcolor[HTML]{ababab}69.34 & \cellcolor[HTML]{b0b0b0}65.70 \\
\textbf{NAPGuard}\cite{NAPGuard} & \cellcolor[HTML]{969696}\underline{87.37} & \cellcolor[HTML]{989898}\textbf{85.23} & \cellcolor[HTML]{9b9b9b}\textbf{83.10} & \cellcolor[HTML]{979797}\underline{85.92} & \cellcolor[HTML]{999999}\textbf{84.14} & \cellcolor[HTML]{9d9d9d}\underline{81.35} \\
\textbf{DIFFender}\cite{Diffender} & \cellcolor[HTML]{ababab}69.41 & \cellcolor[HTML]{aeaeae}67.23 & \cellcolor[HTML]{adadad}68.02 & \cellcolor[HTML]{acacac}68.34 & \cellcolor[HTML]{afafaf}66.22 & \cellcolor[HTML]{b0b0b0}65.73 \\
\textbf{NutNet}~\cite{NutNet} & \cellcolor[HTML]{969696}\textbf{87.42} & \cellcolor[HTML]{9a9a9a}\underline{84.04} & \cellcolor[HTML]{9c9c9c}\underline{82.09} & \cellcolor[HTML]{979797}\textbf{86.12} & \cellcolor[HTML]{9a9a9a}\underline{83.57} & \cellcolor[HTML]{9d9d9d}\textbf{81.42} \\
\textbf{LGS}~\cite{LGS} & \cellcolor[HTML]{a0a0a0}78.37 & \cellcolor[HTML]{a5a5a5}74.81 & \cellcolor[HTML]{a7a7a7}72.63 & \cellcolor[HTML]{a5a5a5}74.23 & \cellcolor[HTML]{a9a9a9}71.48 & \cellcolor[HTML]{aeaeae}67.28 \\
\textbf{Zmask}~\cite{Zmask} & \cellcolor[HTML]{a1a1a1}77.45 & \cellcolor[HTML]{a4a4a4}75.41 & \cellcolor[HTML]{a9a9a9}71.28 & \cellcolor[HTML]{ababab}69.25 & \cellcolor[HTML]{b1b1b1}64.36 & \cellcolor[HTML]{b4b4b4}62.19 \\
\textbf{Jedi}~\cite{Jedi} & \cellcolor[HTML]{acacac}68.37 & \cellcolor[HTML]{b1b1b1}64.19 & \cellcolor[HTML]{b4b4b4}62.40 & \cellcolor[HTML]{ababab}69.63 & \cellcolor[HTML]{afafaf}66.22 & \cellcolor[HTML]{afafaf}66.15 \\ \hline
\textbf{DetectorGuard}\cite{detectorguard} & \cellcolor[HTML]{ababab}68.50 & \cellcolor[HTML]{acacac}68.32 & \cellcolor[HTML]{ffffff}/ & \cellcolor[HTML]{a9a9a9}70.63 & \cellcolor[HTML]{a9a9a9}70.04 & \cellcolor[HTML]{ffffff}/ \\
\textbf{ObjectSeeker}\cite{objectseeker} & \cellcolor[HTML]{ababab}69.16 & \cellcolor[HTML]{acacac}68.91 & \cellcolor[HTML]{ffffff}/ & \cellcolor[HTML]{a9a9a9}71.45 & \cellcolor[HTML]{a9a9a9}71.03 & \cellcolor[HTML]{ffffff}/ \\

\hline
\end{tabular}
}
\caption{\textbf{Comparison of 9 empirical and 2 certified defenses with different numbers of adversarial patches (1, 2, 3) under T-SEA~\cite{T-SEA} attack}. We report person AP@0.5.}
\label{tab:certified}
\end{table}

\subsection{Certified Defenses}
\label{sec:certified}

Due to the deterministic guarantees and rigorous robustness proofs, certified defenses have gained increasing attention~\cite{objectseeker, detectorguard}. However, certified defenses rely on strict threat model assumptions. Therefore, we change the setting to constrained attack scenarios for a fair comparison and evaluate certified and empirical defenses. For example, \cite{objectseeker} requires explicit constraints on the number of patches in a scene and their size specifications. In the default setting, we set the number of vertical/horizontal lines ($k$ = 30) and the filtering threshold ($\tau$ = 0.6). As shown in the Tab.~\ref{tab:certified}, we evaluate defense performance by varying the number of patches. Experimental results indicate that defense effectiveness generally declines slightly as the number of patches increases, though ObjectSeeker~\cite{objectseeker} exhibits relatively smaller performance degradation. However, since ObjectSeeker employs exhaustive enumeration of potential patch locations, its computational cost escalates exponentially with the number of patches, posing a significant barrier to broader adoption. Notably, empirical defenses like PAD and NutNet maintain superior performance (consistently above 80\% AP across all patch counts) compared to certified approaches. Future work could focus on expanding the scope of threat models (e.g., relaxing assumptions on patch counts or sizes) while preserving robustness guarantees, thereby advancing certified defenses from theoretical frameworks toward real-world applicability.

\section{Comprehensive Analyses}
\label{sec:analysis}

\begin{table*}[!t]
\resizebox{\textwidth}{!}{
\begin{tabular}{c | ccc | ccc | ccc | ccc | ccc | ccc}
\hline
\multirow{2}{*}{\textbf{Attack} (w/o defense)} & \multicolumn{3}{c}{\textbf{SAC}~\cite{SAC}} & \multicolumn{3}{c}{\textbf{PAD}~\cite{PAD}} & \multicolumn{3}{c}{\textbf{Adyolo}~\cite{Adyolo}} & \multicolumn{3}{c}{\textbf{NAPGuard}~\cite{NAPGuard}} & \multicolumn{3}{c}{\textbf{DIFFender}~\cite{Diffender}} & \multicolumn{3}{c}{\textbf{NutNet}~\cite{NutNet}} \\ \cline{2-19}
                                  & AP & Diff & SmIoU & AP & Diff & SmIoU & AP & Diff & SmIoU & AP & Diff & SmIoU & AP & Diff & SmIoU & AP & Diff & SmIoU \\
\hline
\textbf{T-SEA}~\cite{T-SEA} (36.57) & \cellcolor[HTML]{ffffff}48.79 & \cellcolor[HTML]{b9b9b9}12.22 & \cellcolor[HTML]{b9b9b9}\underline{21.22} & \cellcolor[HTML]{dddddd}73.55 & \cellcolor[HTML]{969696}\textbf{36.98} & \cellcolor[HTML]{aeaeae}\underline{23.88} & \cellcolor[HTML]{ffffff}56.3 & \cellcolor[HTML]{a7a7a7}19.73 & \cellcolor[HTML]{bfbfbf}\underline{22.39} & \cellcolor[HTML]{a6a6a6}\underline{79.21} & \cellcolor[HTML]{969696}\textbf{42.64} & \cellcolor[HTML]{969696}\textbf{77.18} & \cellcolor[HTML]{ffffff}47.58 & \cellcolor[HTML]{b3b3b3}11.01 & \cellcolor[HTML]{f3f3f3}8.4 & \cellcolor[HTML]{ffffff}68.03 & \cellcolor[HTML]{adadad}31.46 & \cellcolor[HTML]{ededed}18.05\\
\hline
\textbf{TC-EGA}~\cite{TC-EGA} (44.41) & \cellcolor[HTML]{d4d4d4}58.75 & \cellcolor[HTML]{adadad}\underline{14.34} & \cellcolor[HTML]{969696}\textbf{31.84} & \cellcolor[HTML]{e2e2e2}72.78 & \cellcolor[HTML]{b7b7b7}28.37 & \cellcolor[HTML]{d4d4d4}21.38 & \cellcolor[HTML]{d3d3d3}64.32 & \cellcolor[HTML]{a6a6a6}\underline{19.91} & \cellcolor[HTML]{969696}\textbf{35.9} & \cellcolor[HTML]{ffffff}68.2 & \cellcolor[HTML]{c7c7c7}23.79 & \cellcolor[HTML]{c2c2c2}49.23 & \cellcolor[HTML]{c9c9c9}59.98 & \cellcolor[HTML]{9f9f9f}\underline{15.57} & \cellcolor[HTML]{b7b7b7}\underline{11.87} & \cellcolor[HTML]{9f9f9f}80.16 & \cellcolor[HTML]{9e9e9e}\underline{35.75} & \cellcolor[HTML]{969696}\textbf{43.79}\\
\hline
\textbf{AdvPatch}~\cite{Advpatch} (33.89) & \cellcolor[HTML]{efefef}52.37 & \cellcolor[HTML]{969696}\textbf{18.48} & \cellcolor[HTML]{bababa}20.92 & \cellcolor[HTML]{ffffff}68.36 & \cellcolor[HTML]{9f9f9f}\underline{34.47} & \cellcolor[HTML]{d4d4d4}21.35 & \cellcolor[HTML]{f9f9f9}57.34 & \cellcolor[HTML]{969696}\textbf{23.45} & \cellcolor[HTML]{cdcdcd}17.85 & \cellcolor[HTML]{cdcdcd}74.36 & \cellcolor[HTML]{9b9b9b}\underline{40.47} & \cellcolor[HTML]{a8a8a8}\underline{65.18} & \cellcolor[HTML]{ededed}51.59 & \cellcolor[HTML]{969696}\textbf{17.7} & \cellcolor[HTML]{f4f4f4}8.34 & \cellcolor[HTML]{dddddd}72.32 & \cellcolor[HTML]{969696}\textbf{38.43} & \cellcolor[HTML]{e4e4e4}20.8\\
\hline
\textbf{GNAP}~\cite{GNAP} (73.48) & \cellcolor[HTML]{969696}\textbf{73.47} & \cellcolor[HTML]{ffffff}-0.01 & \cellcolor[HTML]{ffffff}0.07 & \cellcolor[HTML]{9f9f9f}\underline{83.15} & \cellcolor[HTML]{ffffff}9.67 & \cellcolor[HTML]{969696}\textbf{25.45} & \cellcolor[HTML]{969696}\textbf{75.48} & \cellcolor[HTML]{fafafa}2.0 & \cellcolor[HTML]{f6f6f6}4.61 & \cellcolor[HTML]{969696}\textbf{81.26} & \cellcolor[HTML]{f1f1f1}7.78 & \cellcolor[HTML]{ffffff}10.59 & \cellcolor[HTML]{a9a9a9}\underline{67.51} & \cellcolor[HTML]{ffffff}-5.97 & \cellcolor[HTML]{969696}\textbf{13.8} & \cellcolor[HTML]{9c9c9c}\underline{80.5} & \cellcolor[HTML]{ffffff}7.02 & \cellcolor[HTML]{a6a6a6}\underline{38.86}\\
\hline
\textbf{DM-NAP}~\cite{DM-NAP} (71.23) & \cellcolor[HTML]{9e9e9e}\underline{71.42} & \cellcolor[HTML]{fdfdfd}0.19 & \cellcolor[HTML]{fafafa}1.32 & \cellcolor[HTML]{969696}\textbf{84.59} & \cellcolor[HTML]{f0f0f0}13.36 & \cellcolor[HTML]{ffffff}18.6 & \cellcolor[HTML]{a8a8a8}\underline{72.16} & \cellcolor[HTML]{ffffff}0.93 & \cellcolor[HTML]{ffffff}1.84 & \cellcolor[HTML]{d0d0d0}74.01 & \cellcolor[HTML]{ffffff}2.78 & \cellcolor[HTML]{d8d8d8}35.01 & \cellcolor[HTML]{969696}\textbf{72.08} & \cellcolor[HTML]{e0e0e0}0.85 & \cellcolor[HTML]{ffffff}7.74 & \cellcolor[HTML]{969696}\textbf{81.32} & \cellcolor[HTML]{f4f4f4}10.09 & \cellcolor[HTML]{ffffff}13.04\\
\hline
\end{tabular}
}
\caption{\textbf{Results for adversarial patches that bypass defenses}. We report AP@0.5 after defenses, the difference in AP@0.5 before and after defenses, and SmIoU. The result without defense is located next to the attack method's name.}
\label{tab:evade}
\end{table*}

In this section, we first investigate the root causes of defense failures in specific adversarial patch instances (Sec~\ref{sec:evade}). Subsequently, we retrain mainstream defense methods using our APDE dataset and demonstrate significant performance improvements (Sec~\ref{sec:retrain}). Finally, we systematically analyze the impact of adversarial patches with varying sizes on defense robustness (Sec~\ref{sec:size}).


\subsection{Defense Failures}
\label{sec:evade}
During the experiments, we observe that the defense methods perform poorly against certain adversarial patches. We select a few representative attack methods and test the defense performance against them, with the results presented in Tab.~\ref{tab:evade}. It is clear that when detectors are attacked by GNAP~\cite{GNAP} and DM-NAP~\cite{DM-NAP} patches, most defense methods show a slight improvement in person AP. In contrast, under attacks from other patches, the defenses are more effective. The main distinction between these patches lies in their visual characteristics, with the former appearing more natural. Previous studies~\cite{NAPGuard,LGS,wang2020high} suggested that the adversarial characteristics of patches are mainly concentrated in the high-frequency components. Unlike non-NAPs, high-frequency components of NAPs appear more similar to their surroundings, making them more deceptive and harder to detect accurately~\cite{NAPGuard}.

\begin{figure}[!t]
    \centering
    \includegraphics[width=\columnwidth]{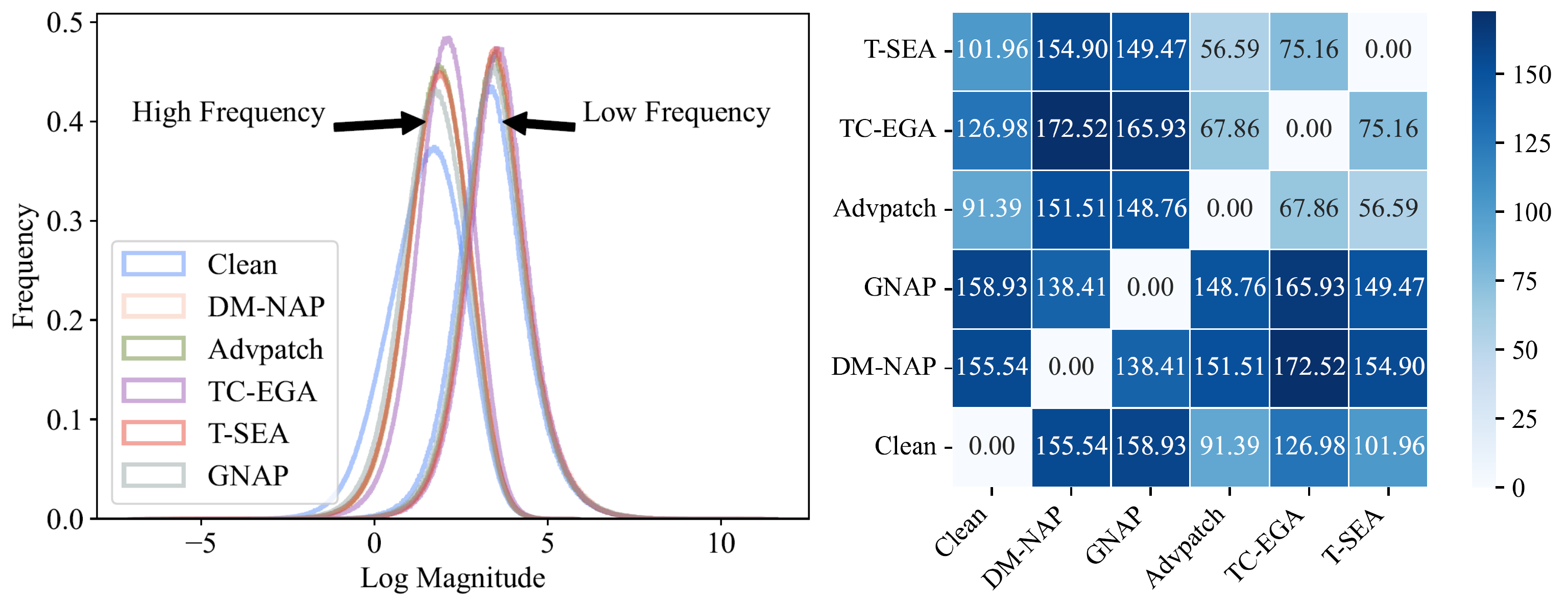}
    \caption{Left: \textbf{Frequency distribution of patches}. Right: \textbf{FID scores between patches} from different attack methods. (In Appendix B.6, we provide additional case studies on frequency distribution and FID scores.)}
    \label{fig:evading_analysis}
\end{figure}

However, our experiments indicate that this explanation is not entirely accurate. As shown in Fig.~\ref{fig:evading_analysis}, we measure the frequency histograms of different adversarial patches and find little difference in the high-frequency components between NAPs and non-NAPs. Compared to background features, their high-frequency components are more similar and difficult to distinguish. Nevertheless, NAPs do evade defenses more effectively than non-NAPs, prompting us to investigate the issue from the data distribution perspective. We compute the Fréchet Inception Distance (FID) scores~\cite{FID} between each of these 5 patches and clean samples. The results in Fig.~\ref{fig:evading_analysis} reveal that non-NAPs exhibit closer FID scores to each other, indicating more similar data distributions, while NAPs are more distantly distributed. Defenses based on patch detection/segmentation or generative models essentially rely on data distribution to determine whether a pixel contains an adversarial patch. \textbf{The difficulty in defending against naturalistic patches lies in the data distribution, rather than the commonly believed high frequencies.}


\begin{table}[!t]
\resizebox{\columnwidth}{!}{
\begin{tabular}{c | cc | cc | cc}
\hline
\multirow{2}{*}{\textbf{Attack}} & \multicolumn{2}{c}{\textbf{SAC}~\cite{SAC}} & \multicolumn{2}{c}{\textbf{Adyolo}~\cite{Adyolo}} & \multicolumn{2}{c}{\textbf{NAPGuard}~\cite{NAPGuard}} \\ \cline{2-7} 
                                             & original & retrained & original & retrained & original & retrained \\ \hline
\textbf{T-SEA}~\cite{T-SEA}                                        & 51.82    & \textbf{71.61}  & 66.61   & \textbf{72.47} & 83.61   & \textbf{86.31} \\ 
\textbf{TC-EGA}~\cite{TC-EGA}                                       & 58.16    & \textbf{71.36}  & 63.49   & \textbf{70.91} & 68.51   & \textbf{85.30} \\ 
\textbf{Advpatch}~\cite{Advpatch}                                    & 56.53    & \textbf{73.29}  & 65.54   & \textbf{72.07} & 78.45   & \textbf{85.10} \\ 
\textbf{GNAP}~\cite{GNAP}                                        & 70.03    & \textbf{76.86}  & 72.94   & \textbf{78.52} & 78.96   & \textbf{85.42} \\ 
\textbf{DM-NAP}~\cite{DM-NAP}                                      & 68.50    & \textbf{76.48}  & 69.26   & \textbf{76.83} & 71.37   & \textbf{85.71} \\ 
\hline
\textbf{AdvCloak}~\cite{Advcloak}                                   & 4.17     & \textbf{71.29}  & 18.29   & \textbf{22.36} & 52.21   & \textbf{73.16} \\ 
\textbf{AdvTshirt}~\cite{AdvTshirt}                                   & 34.27    & \textbf{64.47}  & 8.19    & \textbf{37.53} & 50.21   & \textbf{70.89} \\ 
\hline
\end{tabular}
}
\caption{\textbf{Comparison of defense performance before and after retraining on the APDE dataset}, where AdvCloak and AdvTshirt are out-of-domain patches from the APDE dataset. We report original and retrained AP@0.5 after defenses.}
\label{tab:retrained}
\end{table}

\subsection{Improving Defenses with our APDE dataset}
\label{sec:retrain}
To further validate the above viewpoint, we retrain several defenses using our proposed large-scale APDE dataset (Sec \ref{sec:PDE_dataset}) by splitting into an 6:4 train-test ratio. As shown in Tab.~\ref{tab:retrained}, the retrained defense methods achieve significant improvement in performance (In Appendix B.3, we demonstrate that our APDE dataset more effectively enhances defense performance compared to existing patch datasets). Furthermore, we select two types of patches~\cite{Advcloak,AdvTshirt} that are not included in the training dataset. Many defense methods rely on pre-training strategies where the choice of adversarial patch dataset used during training significantly impacts the defense performance. For example, Adyolo~\cite{Adyolo} improves the overall performance and generalization by alternately optimizing the patch and the defense model through adversarial training. NAPGuard~\cite{NAPGuard} improves detection of naturalistic adversarial examples via Aggressive Feature Aligned Learning. However, despite these advancements, we find that when the dataset covers only a limited range of adversarial patches, the defense method may overfit, leading to poor generalization against out-of-domain patches. \textbf{Our new dataset with diverse patch distributions can be used to improve existing defenses by 15.09\% AP@0.5.}

\begin{figure}[!t]
    \centering
    \includegraphics[width=\columnwidth]{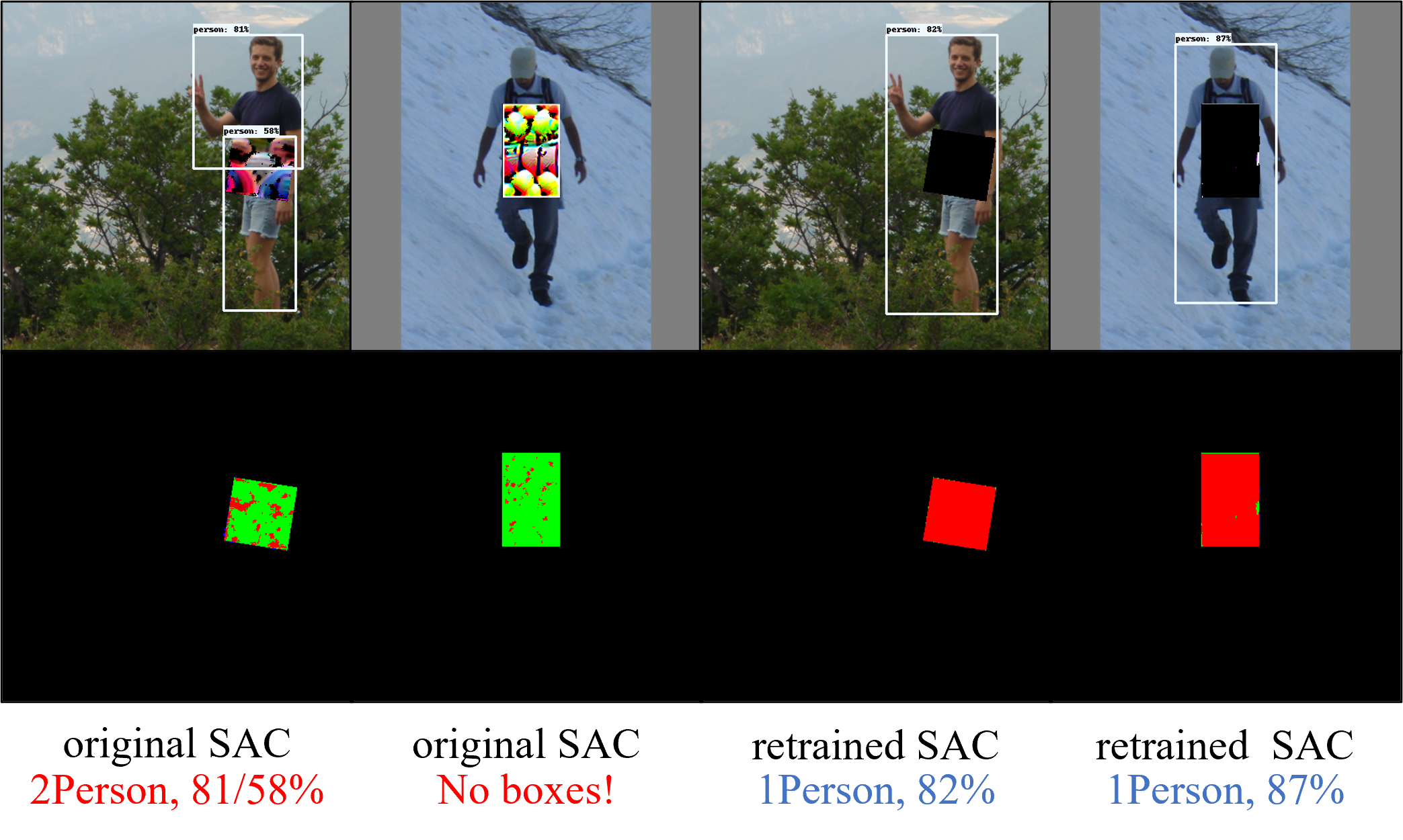}
    \caption{\textbf{Patch detection and defense performance of SAC}~\cite{SAC} \textbf{before and after retraining}. \textbf{Top}: Images with adversarial patches. \textbf{Bottom}: Patch mask images generated by defenses.}
    \label{fig:retrained}
\end{figure}

As shown in Fig.~\ref{fig:retrained}, we take SAC before and after retraining as an example. The first and third columns show T-SEA~\cite{T-SEA} attack patches trained on YOLOv2~\cite{yolov2}. The second and fourth columns show AdvTshirt~\cite{AdvTshirt} attack patches trained on YOLOv2~\cite{yolov2}. The former is included in the training dataset, while the latter is not included. Clearly, the masks generated after retraining more completely eliminate the adversarial patches, restoring normal prediction results.
This is because the APDE dataset contains a wide variety of patches, significantly enriching the defense method's understanding of the patch data distribution. Even for unseen patches, the retrained defenses demonstrate better defense effectiveness and robustness.

\subsection{Impact of Patch Properties}
\label{sec:size}
\begin{figure}[!t]
    \centering
    \includegraphics[width=\columnwidth]{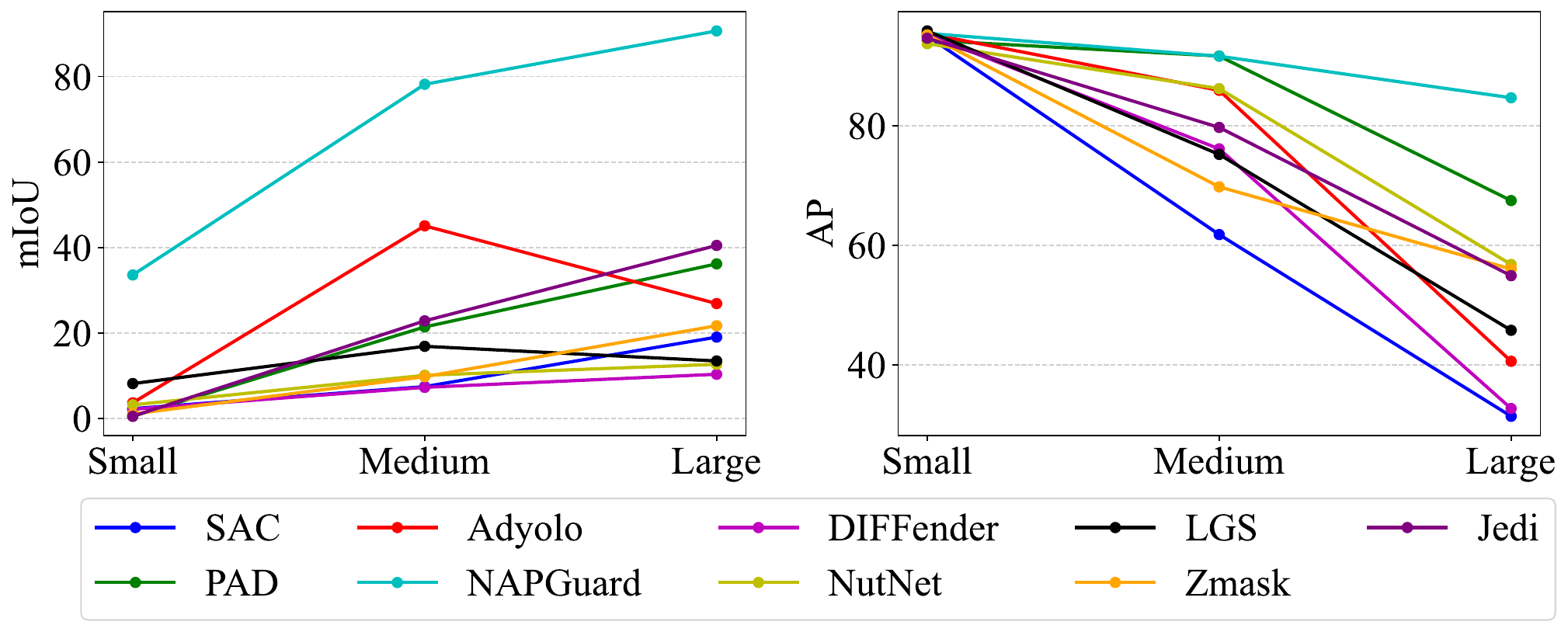}
    \caption{\textbf{Comparison of SmIoU (left) and AP@0.5 (right) for defenses against patches of different sizes.}}
    \label{fig:size1}
\end{figure}

First, we investigate the impact of adversarial patches with varying sizes on defenses. Fig.~\ref{fig:size1} shows the defense and detection performance of each method for adversarial patches of different sizes, measured by person AP and SmIoU. The experimental results reveal that defenses are more successful at detecting larger patches but perform worse in defense performance. When examining the mIoU of SAC~\cite{SAC}, Adyolo~\cite{Adyolo}, NAPGuard~\cite{NAPGuard} and NutNet~\cite{NutNet} in Fig.~\ref{fig:mIoU_asr}, it is obvious that their SmIoU scores are higher than their NmIoU scores, while other defense methods exhibit the opposite phenomenon. We carefully analyze the individual images with patches of different sizes to further explore this issue.

\begin{figure}[!t]
    \centering
    \includegraphics[width=\columnwidth]{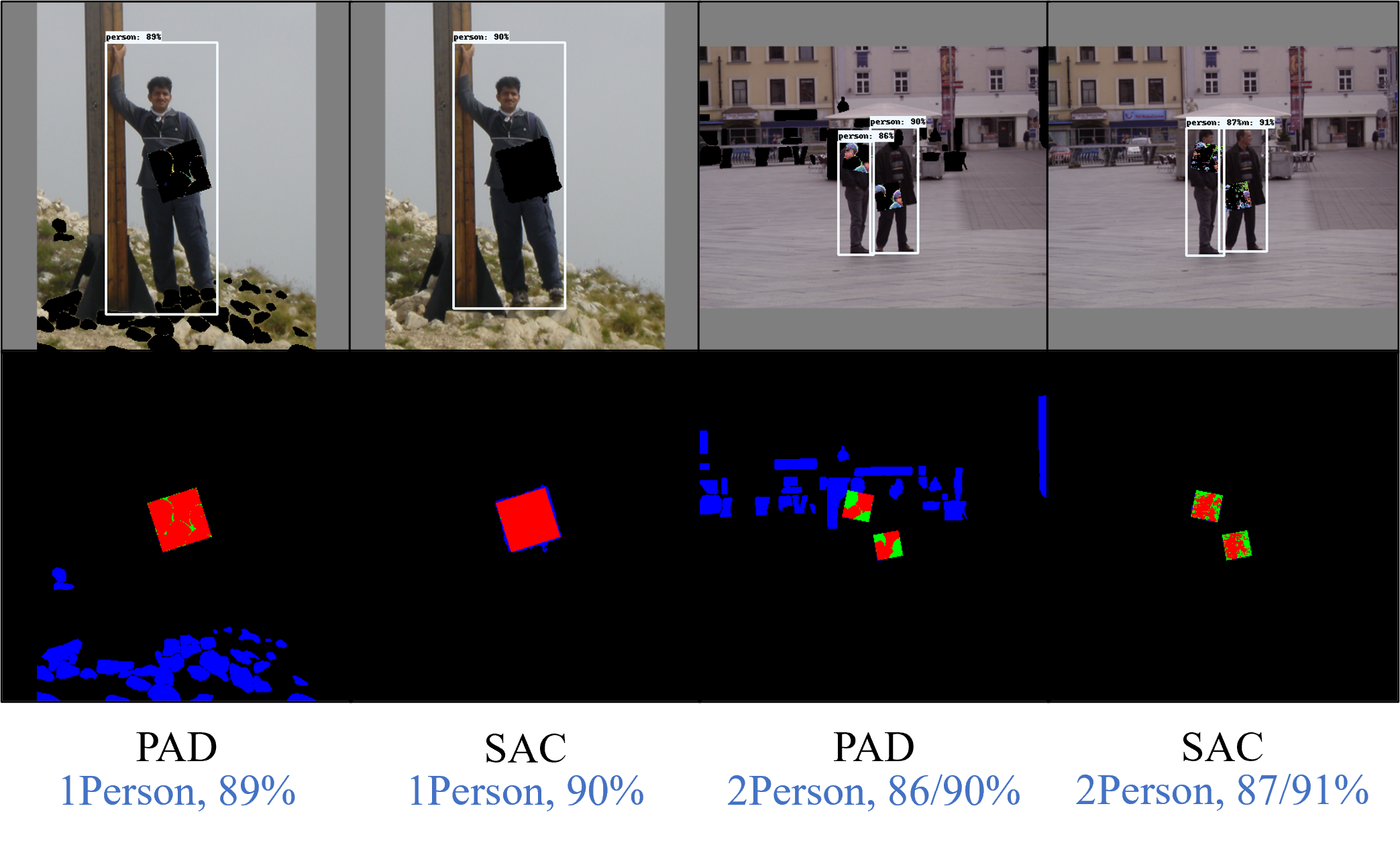}
    \caption{\textbf{Patch detection and defense performance of PAD}~\cite{PAD} \textbf{and SAC}~\cite{SAC} \textbf{for patches of different sizes}. \textbf{Top}: Images with adversarial patches. \textbf{Bottom}: Patch mask images generated by defenses. The first two columns represent large patches, and the last two columns represent small patches.}
    \label{fig:size2}
\end{figure}

As shown in Fig.~\ref{fig:size2}, we display the masks generated by PAD~\cite{PAD} and SAC~\cite{SAC} for both large and small patches. Blue indicates normal background pixels misidentified as patches by the defense. PAD tends to recognize the background as part of the patch, leading to lower mIoU scores. In contrast, SAC tends to generate smaller masks that overlap as much as possible with the patch region, thus benefiting the mIoU calculation. Since larger patches are detected more effectively, SAC’s SmIoU is larger than its NmIoU. On the other hand, PAD’s recognition of large background areas as patches indicates a large number of false positive pixels, resulting in SmIoU being smaller than NmIoU. Therefore, by comparing the SmIoU and NmIoU scores, we can assess whether a defense method tends to misclassify the background as a patch, and use this to evaluate its robustness.

We further explore the impact of patch shapes in Appendix B.2 and patch erasure techniques in Appendix B.4.
We find that while simple patch erasure methods (e.g., black filling) are generally effective, the robustness against irregularly shaped patches varies significantly across methods, with performance degradation often tied to rectangular-based localization or limited training data diversity.

\section{Conclusion}
\label{sec:conclusion}


In this paper, we revisit 11 representative defenses and present the first patch defense benchmark, regarding various metrics. We conduct comprehensive analyses to reveal new insights, including defense vulnerability principles, evaluation metrics, and adaptive attacks. We also construct a new large-scale adversarial patch dataset to evaluate and improve defenses.
For future work, we plan to test defense performance against various physically feasible attacks, such as attacks printed on surfaces like clothes or cars.



\section{Acknowledgments}
\label{sec:acknowledgments}

We would like to thank Yan Zhang, Haoran Fan, and the anonymous reviewers for their valuable feedback. 
This work was supported in part by the National Key Research and Development Program of China under Grant 2023YFB3107401; the National Natural Science Foundation of China (T2341003, 62376210, 62161160337, 62132011, U24B20185, U21B2018, 62206217, 62402377), the Shaanxi Province Key Industry Innovation Program (2023-ZDLGY-38). 
Thanks to the New Cornerstone Science Foundation and the Xplorer Prize. 


{
    \small
    \bibliographystyle{ieeenat_fullname}
    \bibliography{main}
}



\end{document}